\documentclass[runningheads]{llncs}

\usepackage[utf8]{inputenc}
\usepackage[T1]{fontenc}
\usepackage{amsmath,amsfonts,amssymb}
\usepackage{graphicx}
\usepackage{booktabs}
\usepackage{multirow}
\usepackage{xcolor}
\usepackage{algorithm}
\usepackage{algorithmic}
\usepackage{hyperref}
\usepackage{xurl}
\usepackage{eurosym}
\usepackage{tikz}
\usetikzlibrary{arrows.meta,positioning,shapes.geometric,fit,calc,backgrounds}
\usepackage{listings}

\definecolor{codebg}{HTML}{F7F7F7}
\definecolor{promptbg}{HTML}{F0F4FF}
\definecolor{codegreen}{HTML}{2E7D32}
\definecolor{codepurple}{HTML}{7B1FA2}
\definecolor{codeorange}{HTML}{E65100}
\definecolor{codeblue}{HTML}{1565C0}
\definecolor{codegray}{HTML}{757575}
\definecolor{codestring}{HTML}{C62828}
\definecolor{codenumber}{HTML}{00695C}
\definecolor{promptkey}{HTML}{0D47A1}
\definecolor{promptnum}{HTML}{BF360C}
\definecolor{promptcomment}{HTML}{558B2F}
\definecolor{framecolor}{HTML}{BDBDBD}

\lstdefinestyle{pythonstyle}{
  language=Python,
  backgroundcolor=\color{codebg},
  basicstyle=\ttfamily\fontsize{6.5}{7.8}\selectfont,
  keywordstyle=\color{codeblue}\bfseries,
  stringstyle=\color{codestring},
  commentstyle=\color{codegreen}\itshape,
  numberstyle=\color{codegray},
  emphstyle=\color{codepurple},
  emph={SynthesizedCWM,NKState,predict,evaluate,actions,T,bin},
  frame=single,
  rulecolor=\color{framecolor},
  framesep=3pt,
  xleftmargin=4pt,
  xrightmargin=2pt,
  aboveskip=4pt,
  belowskip=2pt,
  showstringspaces=false,
  breaklines=true,
  tabsize=2,
  columns=flexible,
  escapechar=|
}

\lstdefinestyle{promptstyle}{
  backgroundcolor=\color{promptbg},
  basicstyle=\ttfamily\fontsize{6.5}{7.8}\selectfont,
  frame=single,
  rulecolor=\color{framecolor},
  framesep=3pt,
  xleftmargin=4pt,
  xrightmargin=2pt,
  aboveskip=4pt,
  belowskip=2pt,
  showstringspaces=false,
  breaklines=true,
  tabsize=2,
  columns=flexible,
  escapechar=|
}

\newcommand{\CWM}{\textsc{CWM}}
\newcommand{\jump}{\textsc{Jump}}
\newcommand{\onemax}{\textsc{OneMax}}
\newcommand{\lo}{\textsc{LeadingOnes}}
\newcommand{\EAalpha}{\text{EA}_\alpha}

\title{Code World Models for Parameter Control\\in Evolutionary Algorithms}

\author{Camilo Chac\'{o}n Sartori\inst{1} \and 
Guillem Rodr\'{i}guez Corominas\inst{2,3}}

\authorrunning{C. Chac\'{o}n Sartori \and G. Rodr\'{i}guez Corominas}

\institute{ICN2 -- Catalan Institute of Nanoscience and Nanotechnology,
Barcelona, Spain\\
\email{camilo.chacon@icn2.cat} \and
Artificial Intelligence Research Institute (IIIA-CSIC),
Barcelona, Spain \and
Universitat Polit\`{e}cnica de Catalunya (UPC),
Barcelona, Spain\\
\email{guillem.rodriguez.corominas@upc.edu}}

\begin{document}
\maketitle

\begin{abstract}
Can an LLM learn how an optimizer behaves---and use that knowledge to control it?
We extend Code World Models (CWMs), LLM-synthesized Python programs that predict environment dynamics, from deterministic games to stochastic combinatorial optimization. Given suboptimal trajectories of $(1{+}1)$-$\text{RLS}_k$, the LLM synthesizes a simulator of the optimizer's dynamics; greedy planning over this simulator then selects the mutation strength $k$ at each step.
On \lo{} and \onemax{}, CWM-greedy performs within 6\% of the theoretically optimal policy---without ever seeing optimal-policy trajectories.
On \jump{$_k$}, where a deceptive valley causes all adaptive baselines to fail (0\% success rate), CWM-greedy achieves 100\% success rate---without any collection policy using oracle knowledge of the gap parameter.
On the NK-Landscape, where no closed-form model exists, CWM-greedy outperforms all baselines across fifteen independently generated instances ($36.94$ vs.\ $36.32$; $p<0.001$) when the prompt includes empirical transition statistics.
The CWM also outperforms DQN in sample efficiency (200 offline trajectories vs.\ 500 online episodes), success rate (100\% vs.\ 58\%), and generalization ($k{=}3$: 78\% vs.\ 0\%).
Robustness experiments confirm stable synthesis across 5 independent runs.

\end{abstract}

\section{Introduction}
\label{sec:introduction}

Parameter control is a fundamental challenge in evolutionary computation~\cite{karafotias2015parameter,aleti2016systematic}. For the $(1{+}1)$-$\text{RLS}_k$ algorithm, which flips exactly $k$ bits per step, the question reduces to: \emph{which $k$ should we use at each step?} The optimal adaptive policy $k^*(i)$ is known for \lo{} and \onemax{}, but not for complex landscapes such as \jump{$_k$}.

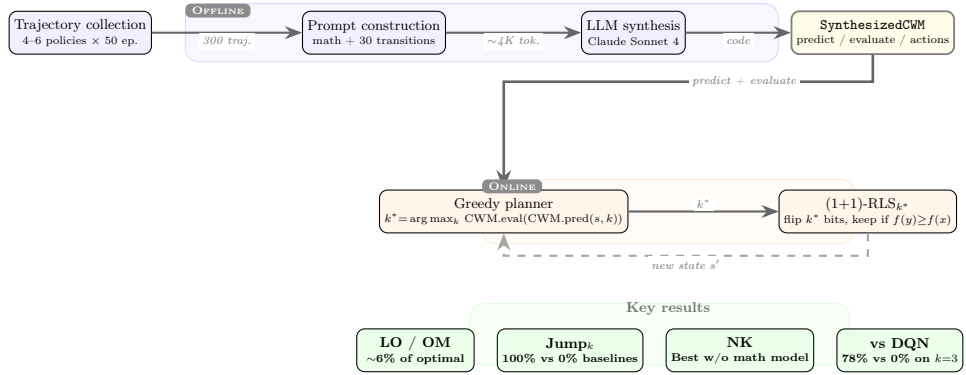
\begin{figure}[t]
\centering
\begin{tikzpicture}[
    scale=0.60, every node/.style={scale=0.60},
    >=Stealth,
    box/.style={draw, rounded corners=3pt, minimum height=0.95cm,
                align=center, font=\small, line width=0.4pt},
    offbox/.style={box, fill=blue!6, minimum width=2.1cm},
    onbox/.style={box, fill=orange!8, minimum width=2.5cm},
    cwmbox/.style={box, fill=yellow!10, draw=black!50, minimum width=2.3cm,
                   line width=0.7pt},
    result/.style={box, fill=green!8, minimum width=2.6cm,
                   font=\small\bfseries},
    arr/.style={->, thick, color=black!60},
    lbl/.style={font=\scriptsize\itshape, text=black!55, fill=white,
                inner sep=2pt},
    phasetag/.style={font=\scriptsize\bfseries\scshape, text=white,
                     fill=black!45, rounded corners=2pt,
                     inner xsep=4pt, inner ysep=2pt},
]

\node[offbox] (traj)
  {Trajectory collection\\[-2pt]{\scriptsize 4--6 policies $\times$ 50 ep.}};
\node[offbox, right=2.0cm of traj] (prompt)
  {Prompt construction\\[-2pt]{\scriptsize math + 30 transitions}};
\node[offbox, right=1.8cm of prompt] (synth)
  {LLM synthesis\\[-2pt]{\scriptsize Claude Sonnet 4}};
\node[cwmbox, right=1.4cm of synth] (cwm)
  {\texttt{SynthesizedCWM}\\[-2pt]{\scriptsize predict / evaluate / actions}};

\draw[arr] (traj) -- node[below, lbl] {300 traj.} (prompt);
\draw[arr] (prompt) -- node[below, lbl] {${\sim}$4K tok.} (synth);
\draw[arr] (synth) -- node[below, lbl] {code} (cwm);

\begin{scope}[on background layer]
  \node[fit=(traj)(prompt)(synth)(cwm),
        fill=blue!3, draw=blue!15, rounded corners=5pt,
        inner xsep=8pt, inner ysep=10pt] (offbg) {};
\end{scope}
\node[phasetag, anchor=north west] at (offbg.north west) {Offline};

\path (prompt.south) -- (synth.south) coordinate[midway] (onanchor);
\node[onbox, below=1.8cm of onanchor, minimum width=3.2cm] (plan)
  {Greedy planner\\[-2pt]{\scriptsize $k^*{=}\arg\max_k$ CWM.eval(CWM.pred($s,k$))}};
\node[onbox, right=2.0cm of plan, minimum width=3.0cm] (rls)
  {$(1{+}1)\text{-RLS}_{k^*}$\\[-2pt]{\scriptsize flip $k^*$ bits, keep if $f(y){\geq}f(x)$}};

\draw[arr, line width=1pt] (cwm.south) -- ++(0,-0.6)
  -| node[pos=0.25, right, lbl] {predict + evaluate} (plan.north);
\draw[arr] (plan) -- node[above, lbl] {$k^*$} (rls);
\draw[arr, dashed, black!35]
  (rls.south) -- ++(0,-0.5) -| node[pos=0.25, below, lbl] {new state $s'$} (plan.south);

\begin{scope}[on background layer]
  \node[fit=(plan)(rls),
        fill=orange!3, draw=orange!15, rounded corners=5pt,
        inner xsep=8pt, inner ysep=12pt] (onbg) {};
\end{scope}
\node[phasetag, anchor=north west] at (onbg.north west) {Online};

\node[font=\small\bfseries, text=black!55, below=0.7cm of onbg.south]
  (rlbl) {Key results};

\node[result, below=0.12cm of rlbl, xshift=-5.6cm] (r1)
  {LO / OM\\[-2pt]{\scriptsize ${\sim}$6\% of optimal}};
\node[result, right=0.3cm of r1] (r2)
  {Jump$_k$\\[-2pt]{\scriptsize 100\% vs 0\% baselines}};
\node[result, right=0.3cm of r2] (r3)
  {NK\\[-2pt]{\scriptsize Best w/o math model}};
\node[result, right=0.3cm of r3] (r4)
  {vs DQN\\[-2pt]{\scriptsize 78\% vs 0\% on $k{=}3$}};

\begin{scope}[on background layer]
  \node[fit=(rlbl)(r1)(r2)(r3)(r4),
        fill=green!3, draw=green!12, rounded corners=5pt,
        inner xsep=5pt, inner ysep=5pt] {};
\end{scope}

\end{tikzpicture}
\caption{Overview of the CWM approach. \textbf{Offline}: trajectory data and a problem description are assembled into a prompt; an LLM synthesizes a Python world model in a single API call (${\sim}$\euro\,0.04). \textbf{Online}: the greedy planner queries the CWM at each step to select $k^*$. \textbf{Bottom}: key results across four benchmarks.}
\label{fig:overview}
\end{figure}

Existing \emph{adaptive} approaches---$\EAalpha{}$~\cite{doerr2018ea_alpha} and self-adjusting mechanisms~\cite{doerr2019self_adjusting}---adjust $k$ via multiplicative update rules ($k \to A \cdot k$ on improvement, $k \to A^{-1/b} \cdot k$ otherwise) that require no problem-specific knowledge. These rules have runtime guarantees on unimodal landscapes but decrease $k$ during stagnation, which is counterproductive on deceptive landscapes.
We pursue a different approach based on Code World Models (\CWM{}s)~\cite{dainese2024cwm}: an LLM synthesizes a compact Python program that predicts how the optimizer state evolves under different parameter choices, enabling greedy planning.
\CWM{}s were originally proposed for deterministic game environments; we extend them to stochastic combinatorial optimization, where the world model must encode probabilistic transitions (Figure~\ref{fig:overview}).

Our contributions:
\begin{enumerate}
    \item We extend CWMs to stochastic combinatorial optimization and show that greedy planning suffices (vs.\ Monte Carlo Tree Search (MCTS) in the original work).
    \item On \lo{}, CWM-greedy achieves 1.06$\times$ optimal, outperforming all adaptive baselines ($p < 0.0001$). On \onemax{}, CWM performs within 2\% of optimal, though comparable to \texttt{RLS\_1}. In both cases, the training trajectories contain \emph{no oracle policies}---the LLM infers the correct strategy from problem structure alone.
    \item On \jump{$_k$}, CWM-greedy achieves 100\% success rate where all adaptive baselines fail---without any collection policy using $k_\text{jump}$---the main result of this paper.
    \item On \jump{$_k$} and the NK-Landscape, enriching the prompt with empirical transition statistics enables CWM to outperform all baselines---demonstrating that structured trajectory summaries can substitute for closed-form models and oracle knowledge.
    \item CWM outperforms DQN in sample efficiency (200 offline trajectories vs.\ 500 online episodes), success rate (100\% vs.\ 58\%), and generalization ($k{=}3$: 78\% vs.\ 0\%).
\end{enumerate}

\paragraph{Related work.}
Adaptive parameter control spans from Rechenberg's 1/5-rule~\cite{rechenberg1973} to $\EAalpha{}$~\cite{doerr2018ea_alpha} and modern self-adjusting mechanisms~\cite{doerr2019self_adjusting}; see~\cite{karafotias2015parameter,aleti2016systematic} for surveys.
\jump{$_k$}~\cite{jansen2002analysis} is a canonical deceptive benchmark. Doerr et al.~\cite{doerr2017fast} showed that heavy-tailed mutation achieves $O(n^k / k^{k-1})$ without knowing $k$; Friedrich et al.~\cite{friedrich2018heavy} formalized this for single-objective optimization. Opris et al.~\cite{opris2024tight} proved tight $O(4^k/p_c)$ bounds for $(\mu{+}1)$~GA with crossover. No closed-form optimal \emph{adaptive} policy for $(1{+}1)$-RLS$_k$ on Jump$_k$ is known; our CWM synthesizes a greedy policy that achieves 100\% success rate.

On the learning side, Eiben et al.~\cite{eiben2007rl_ea} introduced online RL for EA control; Sharma et al.~\cite{sharma2019deep} applied Deep Double DQN to Differential Evolution parameter control. Our comparison shows that DQN struggles on deceptive landscapes due to exploration limitations.
FunSearch~\cite{romera2024funsearch} demonstrated LLMs encoding combinatorial reasoning into code; we use the LLM for a \emph{world model} rather than the solution itself.
CWMs~\cite{dainese2024cwm} replace neural world models~\cite{ha2018world,schrittwieser2020muzero} with LLM-synthesized programs; we extend them from deterministic games to stochastic optimization, revealing that greedy planning suffices and MCTS overhead is unnecessary in Markovian settings.

The paper unfolds as follows. Section~\ref{sec:method} presents the CWM method for parameter control. Section~\ref{sec:setup} describes the experimental setup, Section~\ref{sec:results} the results, and Section~\ref{sec:discussion} the discussion. We conclude with a pragmatic justification for using LLMs not as a replacement for formal analysis, but as a complement to it.

\section{Method: Applying CWMs to Parameter Control}
\label{sec:method}

\CWM{}s~\cite{dainese2024cwm} are LLM-synthesized Python programs that act as environment simulators. We apply them to adaptive parameter control in $(1{+}1)$-$\text{RLS}_k$---a mutation-only optimizer that maintains a bitstring $x \in \{0,1\}^n$ and, at each step, flips exactly $k$\footnote{Following standard conventions~\cite{jansen2002analysis,doerr2017fast}, $k$ denotes both the mutation strength of $(1{+}1)$-$\text{RLS}_k$ (a variable) and the gap parameter of \jump{$_k$} (a fixed benchmark constant); the meaning is clear from context. The NK epistasis parameter is denoted $K$ (uppercase).} uniformly random bits, accepting the offspring if $f(y) \geq f(x)$.\footnote{This differs from the $(1{+}1)$-EA, which flips each bit independently with probability $k/n$; RLS$_k$ always flips \emph{exactly} $k$ bits.}
The central question is: \emph{which $k$ should we use at each step?}
Rather than designing adaptive rules by hand, we let the LLM synthesize a simulator of the optimizer's dynamics---a CWM---and then use it to choose the best $k$ via one-step lookahead.

The pipeline has three stages:

\textbf{1. Trajectory collection.}
We run $(1{+}1)$-$\text{RLS}_k$ with diverse policies for 50 runs each ($n=50$), yielding 200--300 trajectories with full transition records.
Crucially, \emph{none} of the collection policies uses the optimal $k^*(i)$---the CWM must infer good actions from suboptimal demonstrations alone:
\begin{itemize}\setlength{\itemsep}{0pt}\setlength{\parskip}{0pt}
\item \emph{Problem-agnostic} (all benchmarks): random, fixed-$1$, $\sqrt{n}$, decreasing ($k{=}\max(1,n{-}i)$).
\item \emph{Problem-specific} (\lo{}/\onemax{} only): fixed-$\lfloor n/2 \rfloor$, $k{=}i{+}1$.
\end{itemize}
\vspace{-2pt}
For \jump{$_k$} and NK, only the four agnostic policies are used---no collection policy uses $k_\text{jump}$ or any other oracle knowledge.
The synthesis prompt includes the mathematical problem description plus 30 uniformly sampled transitions (stratified across fitness levels). For \jump{$_k$} and NK---where dynamics are harder to infer from mathematics or no closed-form model exists---the prompt is \emph{enriched} with an empirical transition table summarizing $P(\text{improve} \mid \text{fitness}, k)$ and mean $\Delta f$ per (fitness range, $k$) bin.

\textbf{2. CWM synthesis.}
An LLM (Claude Sonnet 4, \texttt{claude-sonnet-4-20250514}) receives the problem description and trajectory samples, then generates a \texttt{SynthesizedCWM} class with four methods: \texttt{predict\_next\_state}, \texttt{get\_legal\_actions}, \texttt{evaluate\_state}, and \texttt{is\_terminal}.
A key design insight is the \texttt{normalized\_fitness} trick: the continuous expected fitness (not just the integer) allows discrimination between actions with different improvement probabilities---analogous to maximizing \emph{drift} $E[f(y) - f(x) \mid f(x) = i, k]$ per step, which is known to yield near-optimal policies for \lo{}~\cite{doerr2019self_adjusting}.
Generated code undergoes automated validation with refinement (up to 5 attempts). Each synthesis call uses ${\sim}4{,}000$ input tokens and ${\sim}2{,}000$ output tokens; at current API pricing, a single CWM costs ${\sim}$\euro 0.04 to synthesize.

\textbf{3. CWM-greedy planning.}
At each step, one-step lookahead selects $k^* = \arg\max_{k \in \mathcal{K}(s)} \texttt{evaluate\_state}(\texttt{predict\_next\_state}(s, k))$.

Algorithm~\ref{alg:cwm} summarizes the pipeline.

\begin{algorithm}[t]
\caption{\CWM{}-greedy for adaptive parameter control}
\label{alg:cwm}
\begin{algorithmic}[1]
\STATE \textbf{Offline:} Run $(1{+}1)$-$\text{RLS}_k$ with diverse policies; collect $\mathcal{D}$
\STATE \quad Prompt LLM with problem description $+$ samples from $\mathcal{D}$
\STATE \quad Extract and validate \texttt{SynthesizedCWM}
\STATE \textbf{Online:} Initialize random $x \in \{0,1\}^n$
\WHILE{not terminal}
    \FOR{each $k \in$ \texttt{CWM.get\_legal\_actions}$(s)$}
        \STATE $v_k \leftarrow$ \texttt{CWM.evaluate\_state}$($\texttt{CWM.predict\_next\_state}$(s, k))$
    \ENDFOR
    \STATE Apply $(1{+}1)$-$\text{RLS}_{k^*}$ with $k^* = \arg\max_k v_k$
\ENDWHILE
\end{algorithmic}
\end{algorithm}

Table~\ref{tab:cwm_comparison} contrasts our extension with the original CWM framework.

\begin{table}[t]
\centering
\caption{Original CWM framework~\cite{dainese2024cwm} vs.\ this work.}
\label{tab:cwm_comparison}
\small
\begin{tabular}{@{}lll@{}}
\toprule
\textbf{Aspect} & \textbf{Original CWM} & \textbf{This work} \\
\midrule
Environments  & 18 Gymnasium games        & 4 EA benchmarks \\
Dynamics      & Deterministic             & Stochastic combinatorial \\
LLM input     & Env.\ docs + trajectories & Math / empirical stats + traj. \\
Synthesis     & GIF-MCTS (iterative)      & Single-shot + refinement \\
Planning      & MCTS (5{,}000 rollouts)   & Greedy 1-step lookahead \\
Baselines     & Model-free RL             & Adaptive rules \&  DQN \\
Key result    & Sample efficiency          & 100\% vs.\ 0\% on deceptive landscape \\
\bottomrule
\end{tabular}
\end{table}

\section{Experimental Setup}
\label{sec:setup}

All experiments use $n = 50$.
Table~\ref{tab:benchmarks} summarizes the four benchmarks in order of increasing difficulty and shows how CWM operates on each.

\begin{table}[t]
\centering
\caption{Benchmark progression: each problem adds a new challenge for CWM-based parameter control.}
\label{tab:benchmarks}
\small
\resizebox{\textwidth}{!}{%
\begin{tabular}{@{}lllll@{}}
\toprule
& \textbf{\lo{}} & \textbf{\onemax{}} & \textbf{\jump{$_k$}} & \textbf{NK} \\
\midrule
Landscape        & Unimodal        & Unimodal           & Deceptive valley    & Rugged / epistatic \\
Optimal $k^*$    & Known (closed)  & Known (closed)     & Unknown             & Unknown \\
Key challenge    & Smooth $k$ ramp & Sharp policy cliff  & Valley crossing     & No math model \\
CWM input        & Math description & Math description   & Math + empirical table & Empirical table \\
Budget           & $0.8n^2{=}2{,}000$ & $5n\ln n{=}978$    & $10{,}000$          & $10{,}000$ \\
What CWM learns  & $k \downarrow$ as $i \uparrow$ & Cliff at $n/2$ & $k \uparrow$ at edge & Transition lookup \\
\bottomrule
\end{tabular}}
\end{table}

For \jump{$_k$}: $k = 2$ (expected valley crossing time from the edge using RLS$_k$ with $k{=}2$: $\binom{50}{2} = 1{,}225$ steps).
For NK: $K = 2$, fifteen independently generated landscape instances; trajectories are collected and the CWM is synthesized on the first instance only, then evaluated on all fifteen without re-training.
Per-episode budgets are set proportional to the theoretical expected runtimes: $0.8\,n^2 = 2{,}000$ for \lo{} (from $\Theta(n^2)$), $5\,n\ln n = 978$ for \onemax{} (from $\Theta(n\ln n)$), and $10{,}000$ for \jump{$_k$} (${\approx}8\times$ the expected valley crossing time $\binom{50}{2} = 1{,}225$) and NK.
All main evaluations use 100 independent runs; generalization experiments use 50 runs.
Evaluation episodes use matched seeds across policies (seed $= 31 \cdot i$ for episode $i$), so all policies face the same initial bitstrings in each run.
Policies compared: \texttt{optimal} (where available), \texttt{CWM-greedy}, adaptive multiplicative rules ($\EAalpha(2,0.5)$, $\EAalpha(1.3,0.75)$, \texttt{fifth\_rule}~\cite{rechenberg1973}, \texttt{self\_adjusting}~\cite{doerr2019self_adjusting}), problem-specific heuristics (\texttt{stagnation}, \texttt{RLS\_jump\_k}---both require $k_\text{jump}$ as input), and fixed/random baselines (\texttt{RLS\_1}, \texttt{random\_k}).
Tables report the most informative subset per benchmark.
Statistical significance is assessed with the two-sided Mann-Whitney U test ($\alpha = 0.05$) for single-benchmark comparisons (\lo{}, \onemax{}, \jump{$_k$}), where each run is independent. For NK, where runs are paired by instance seed, we use a paired $t$-test over per-instance means (15 pairs).
DQN baseline\footnote{Two hidden layers of 128 units, ReLU, experience replay (50K buffer), target network updates every 500 steps, $\epsilon$-greedy $1.0 \to 0.05$. State: normalized fitness, progress, region, stagnation counter~\cite{sharma2019deep}. Training wall time: ${\sim}26$ min for 500 episodes on CPU, vs.\ ${\sim}30$\,s for a single CWM synthesis call.} trained with 500 online episodes for comparison.

\section{Results}
\label{sec:results}

\subsection{Unimodal Benchmarks: LeadingOnes \& OneMax}
\label{sec:results_unimodal}

We begin with two benchmarks whose optimal policies are known in closed form, providing a ground truth against which to validate the CWM.
$\lo(x) = \max\{i : x_1 = \cdots = x_i = 1\}$ has optimal policy $k^*(i) = \lfloor n/(i+1) \rfloor$---a smooth descending ramp.
$\onemax(x) = |x|_1$ has optimal policy with a sharp cliff at $i \approx n/2$, switching abruptly from large to small $k$.
The question is: can the CWM recover these policies from non-optimal trajectories alone?

Table~\ref{tab:unimodal_results} presents the results.
On \lo{}, \CWM{}-greedy achieves 1{,}045 steps, within 6\% of optimal (983), significantly outperforming all baselines ($p < 0.0001$).
On \onemax{}, CWM-greedy performs within 2\% of optimal (190 vs.\ 186), comparable to \texttt{RLS\_1} (189)---on this easier benchmark, all policies perform similarly.
CWM-MCTS underperforms CWM-greedy on \lo{} (1{,}273 vs.\ 1{,}045) and matches RLS$_1$ exactly---MCTS rollouts collapse to $k{=}1$ at every state, failing to exploit the CWM's score gradients. The parameter control problem is effectively Markovian with horizon~1, so MCTS adds $100\times$ overhead (5{,}000 CWM calls vs.\ 50) with no benefit.

\begin{table}[t]
\centering
\caption{Unimodal benchmark results (100 runs each). \lo{}: $n{=}50$, budget${=}2{,}000$. \onemax{}: $n{=}50$, budget${=}978$.}
\label{tab:unimodal_results}
\small
\begin{tabular}{@{}llrrr@{}}
\toprule
\textbf{Benchmark} & \textbf{Policy} & \textbf{Mean} & \textbf{Std} & \textbf{vs opt.} \\
\midrule
\multirow{5}{*}{\lo{}}
& \texttt{optimal}           &  983 & 259 & 1.00$\times$ \\
& \textbf{CWM-greedy}        & \textbf{1{,}045} & \textbf{248} & \textbf{1.06$\times$} \\
& \texttt{fifth\_rule}       & 1{,}256 & 299 & 1.28$\times$ \\
& \texttt{CWM-MCTS}          & 1{,}273 & 316 & 1.29$\times$ \\
& \texttt{RLS\_1}            & 1{,}273 & 316 & 1.29$\times$ \\
\midrule
\multirow{4}{*}{\onemax{}}
& \texttt{optimal}           & 186 &  67 & 1.00$\times$ \\
& \texttt{RLS\_1}            & 189 &  58 & 1.02$\times$ \\
& \textbf{CWM-greedy}        & \textbf{190} & \textbf{60} & \textbf{1.02$\times$} \\
& \texttt{self\_adjusting}   & 196 &  61 & 1.05$\times$ \\
\bottomrule
\end{tabular}
\end{table}

\begin{figure}[t]
    \centering
    \includegraphics[width=0.65\textwidth]{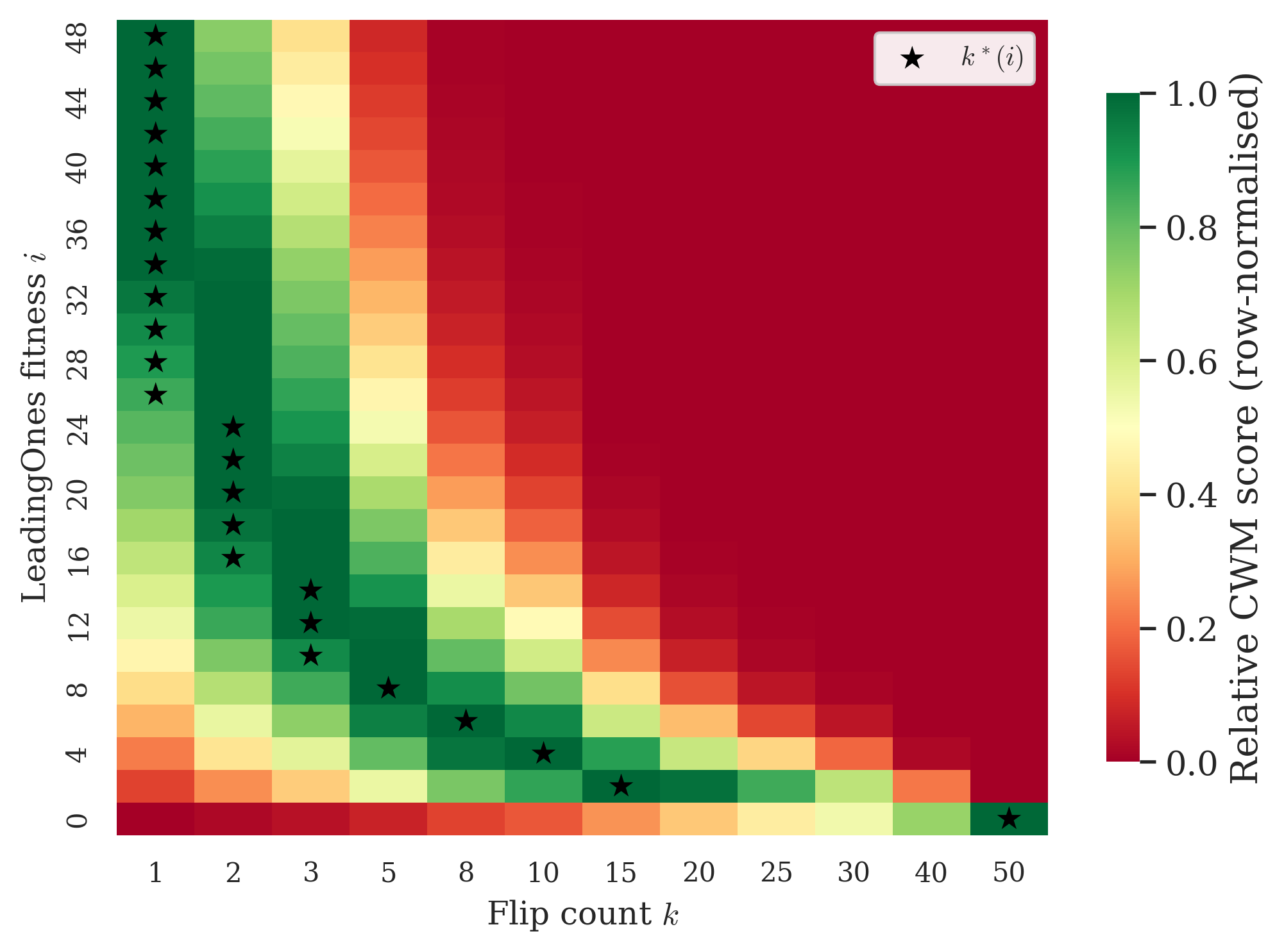}
    \caption{\lo{}: CWM score heatmap over (fitness $i$, parameter $k$). Black stars ($\star$) mark the optimal $k^*(i) = \lfloor n/(i+1) \rfloor$, snapped to the nearest column when $k^*$ is not in the displayed set. Stars appear only at $k$ values that are optimal for at least one fitness level; columns without stars (e.g.\ $k{=}20, 25, 30, 40$) are never greedy-optimal for any $i$.}
    \label{fig:lo_heatmap}
\end{figure}

The CWM achieves high policy correlation on both benchmarks: \lo{} $\tau = 0.784$, $\rho = 0.916$; \onemax{} $\rho = 0.939$, 82\% exact match. On \onemax{}, the CWM captures the sharp cliff in the optimal policy: the policy switches abruptly from $k \gg 1$ (far from optimum) to $k{=}1$ (close), a discontinuity that simple adaptive rules cannot track.
Figure~\ref{fig:lo_heatmap} visualizes the \lo{} CWM's implicit policy: the descending ramp $k^*(i) \approx \lfloor n/(i+1) \rfloor$ is clearly visible, closely matching the theoretically optimal policy (black stars).

\subsection{Jump$_k$: The Deceptive Valley}
\label{sec:results_jk}

The unimodal benchmarks reward steady progress; \jump{$_k$} punishes it. Near the optimum, fitness \emph{drops} into a deceptive valley: $\jump_k(x) = k + |x|_1$ in the normal region ($|x|_1 \leq n{-}k$), but decreases when $n{-}k < |x|_1 < n$~\cite{jansen2002analysis}. Escaping from the valley edge requires flipping exactly $k$ specific zero-bits in one step; with RLS$_k$ this has probability $\binom{n}{k}^{-1}$, giving expected waiting time $\binom{n}{k} = \Theta(n^k / k!)$~\cite{droste2002analysis}. No closed-form optimal \emph{adaptive} policy is known.

This makes \jump{$_k$} the critical benchmark: adaptive baselines designed for unimodal functions see stagnation and decrease $k$---exactly the wrong response.

Unlike \lo{} and \onemax{}, where the mathematical description alone suffices, the Jump$_k$ synthesis prompt is enriched with an empirical transition table---analogous to the NK approach (Section~\ref{sec:results_nk})---because valley-crossing dynamics are difficult to infer from the mathematical definition alone. Crucially, the training trajectories use only the four problem-agnostic policies (random, fixed-$1$, $\sqrt{n}$, decreasing): no collection policy uses $k_\text{jump}$.

Table~\ref{tab:jk_results} presents the results.
CWM-greedy achieves 100\% success rate with mean 1{,}342 steps---comparable to the greedy-optimal policy (1{,}346).
The \texttt{greedy-opt} policy is computed via exact combinatorics: for each fitness level $i$, we enumerate all $k$ and compute the expected fitness gain using the hypergeometric distribution over bit-flip outcomes, selecting $k^*(i) = \arg\max_k E[\Delta f \mid f(x) = i, k]$. This is a 1-step greedy optimum, not a full MDP solution.
All adaptive baselines fail: both $\EAalpha{}$ configurations achieve 0\% success rate.

\begin{table}[t]
\centering
\caption{\jump{$_k$} results ($n=50$, $k=2$, budget${}=10{,}000$, 100 runs).}
\label{tab:jk_results}
\begin{tabular}{@{}lrrrr@{}}
\toprule
\textbf{Policy} & \textbf{Mean} & \textbf{Std} & \textbf{Median} & \textbf{SR\%} \\
\midrule
\texttt{greedy-opt}         & 1{,}346 & 1{,}311 & 940 & 100\% \\
\textbf{CWM-greedy}        & \textbf{1{,}342} & \textbf{1{,}328} & \textbf{1{,}003} & \textbf{100\%} \\
\texttt{stagnation}        & 1{,}506 & 1{,}427     & 1{,}194     & 99\% \\
\texttt{RLS\_jump\_k}      & 5{,}226 & 4{,}176 & 2{,}820 & 58\% \\
$\EAalpha(2, 0.5)$         & 10{,}000 &    0   & 10{,}000 & 0\% \\
$\EAalpha(1.3, 0.75)$      & 10{,}000 &    0   & 10{,}000 & 0\% \\
\texttt{RLS\_1}            & 10{,}000 &    0   & 10{,}000 & 0\% \\
\bottomrule
\end{tabular}
\end{table}

\paragraph{Why adaptive baselines fail.}
At the valley edge (fitness $= n$), the only improvement requires flipping exactly $k$ zero-bits simultaneously. Since no improvement occurs, $\EAalpha{}$ \emph{decreases} $k$---the wrong response---converging to $k=1$, from which escape is impossible.

\begin{figure}[t]
    \centering
    \includegraphics[width=0.65\textwidth]{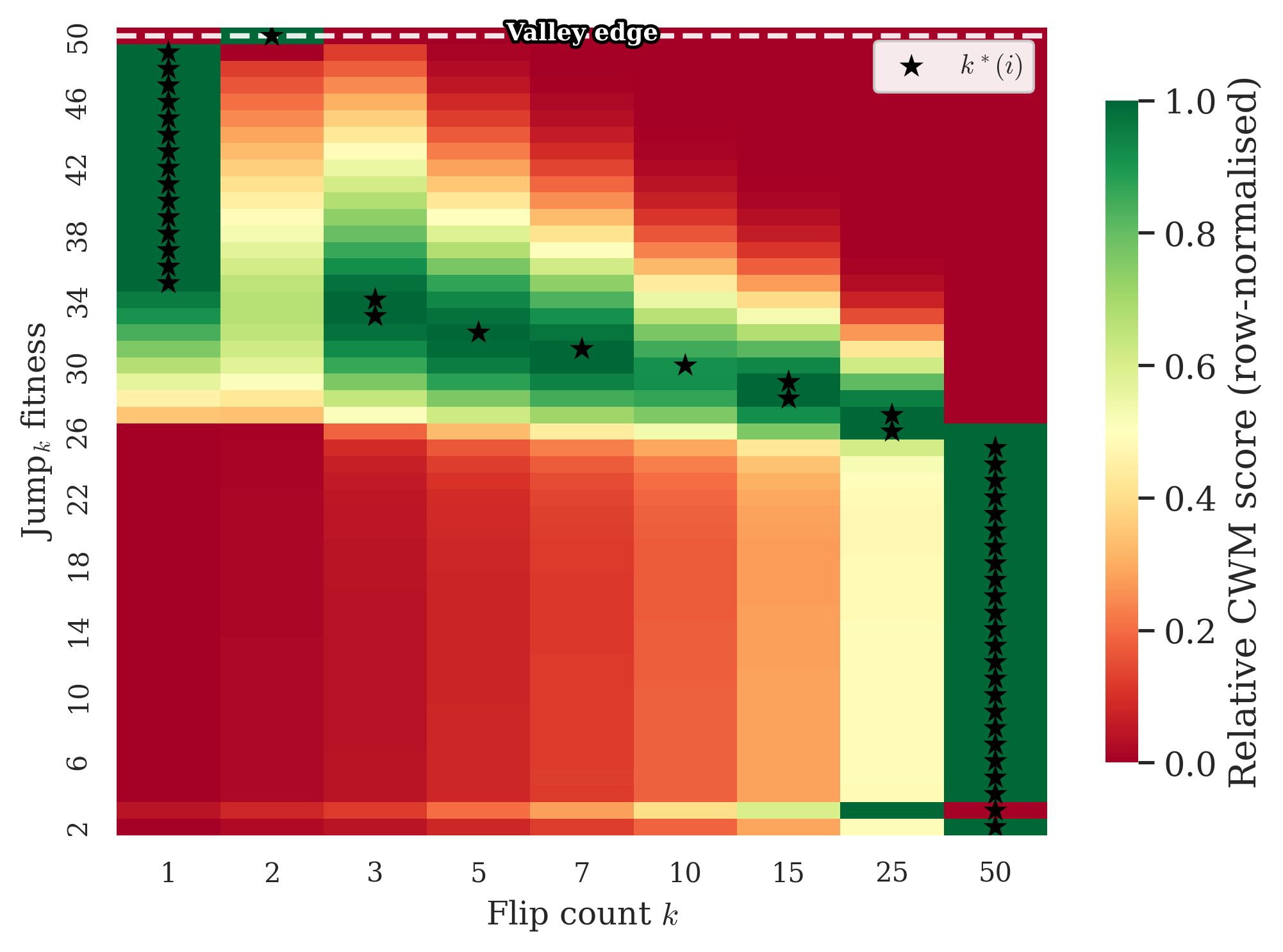}
    \caption{\jump{$_k$} ($k{=}2$): CWM score heatmap over (fitness, parameter $k$). Black stars ($\star$) mark the greedy-optimal $k^*(i)$, snapped to the nearest displayed column. At the valley edge (fitness $= n$), the CWM correctly predicts that only $k{=}2$ leads to improvement---opposite to all adaptive baselines, which decrease $k$ during stagnation.}
    \label{fig:jk_heatmap}
\end{figure}

\paragraph{Why CWM-greedy succeeds.}
The synthesized CWM combines the mathematical hypergeometric model with empirical evidence: at the valley edge, only $k = k_\text{jump}$ yields nonzero improvement probability, achieving valley-edge $\tau = 1.0$---perfect action ranking at the critical decision point (Figure~\ref{fig:jk_heatmap}). The overall correlation is high ($\tau = 0.898$, $\rho = 0.948$), confirming the CWM captures valley dynamics accurately.

\subsection{Comparison with DQN}
\label{sec:results_rl}

DQN~\cite{mnih2015human} has been applied to parameter control in differential evolution~\cite{sharma2019deep}; we use it as our reinforcement learning (RL) baseline.
Table~\ref{tab:dqn_results} compares CWM against DQN on \jump{$_k$}.

\begin{table}[t]
\centering
\caption{\CWM{} vs.\ DQN on \jump{$_k$} ($n=50$, $k=2$, 100 eval.\ runs).}
\label{tab:dqn_results}
\begin{tabular}{@{}lrrl@{}}
\toprule
\textbf{Policy} & \textbf{Mean} & \textbf{SR\%} & \textbf{Data} \\
\midrule
\texttt{optimal}           & 1{,}346 & 100\% & --- \\
\textbf{CWM-greedy}        & \textbf{1{,}342} & \textbf{100\%} & \textbf{200 offline traj.} \\
\texttt{stagnation}        & 1{,}506 & 99\% & --- (heuristic) \\
\texttt{DQN} (500 ep.)     & 5{,}210 &  58\% & 500 online ep. \\
\texttt{DQN} (200 ep.)     & 10{,}000 &   0\% & 200 online ep. \\
\bottomrule
\end{tabular}
\end{table}

\begin{figure}[t]
    \centering
    \begin{minipage}[t]{0.48\textwidth}
        \centering
        \includegraphics[width=\textwidth]{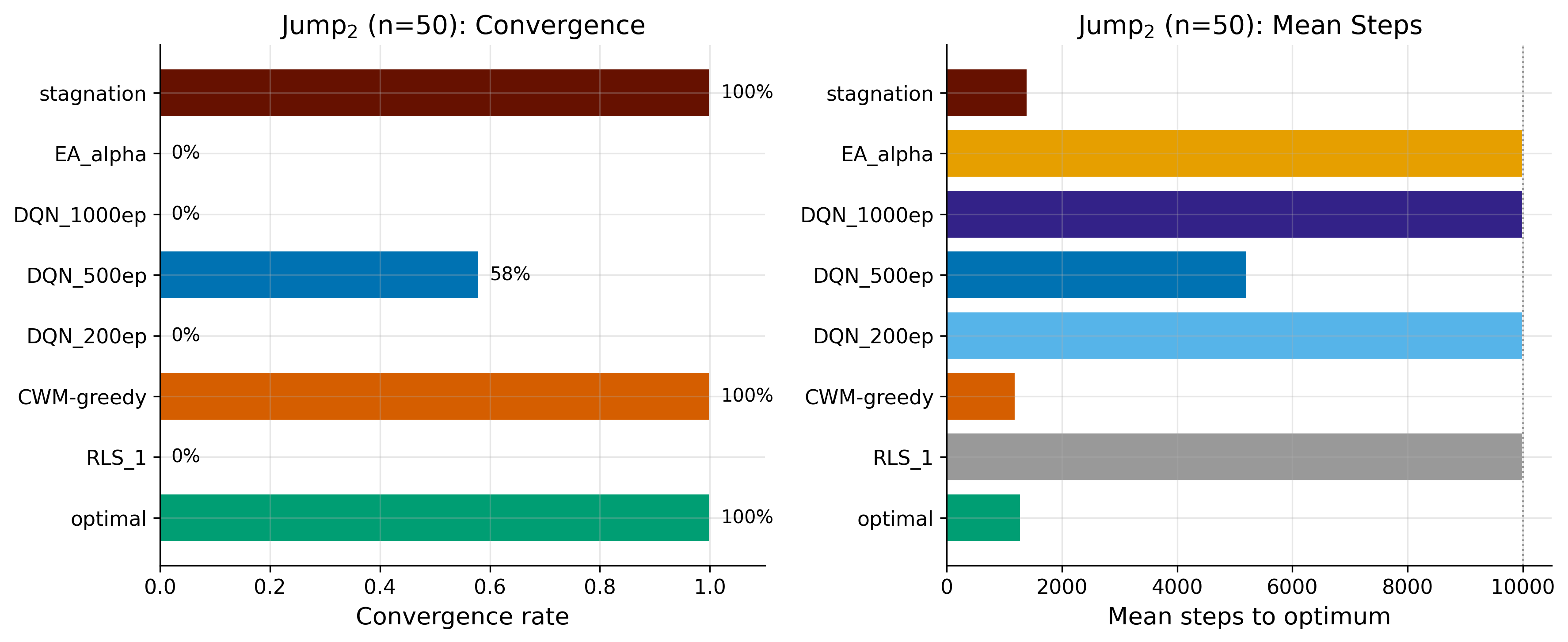}
    \end{minipage}%
    \hfill
    \begin{minipage}[t]{0.48\textwidth}
        \centering
        \includegraphics[width=\textwidth]{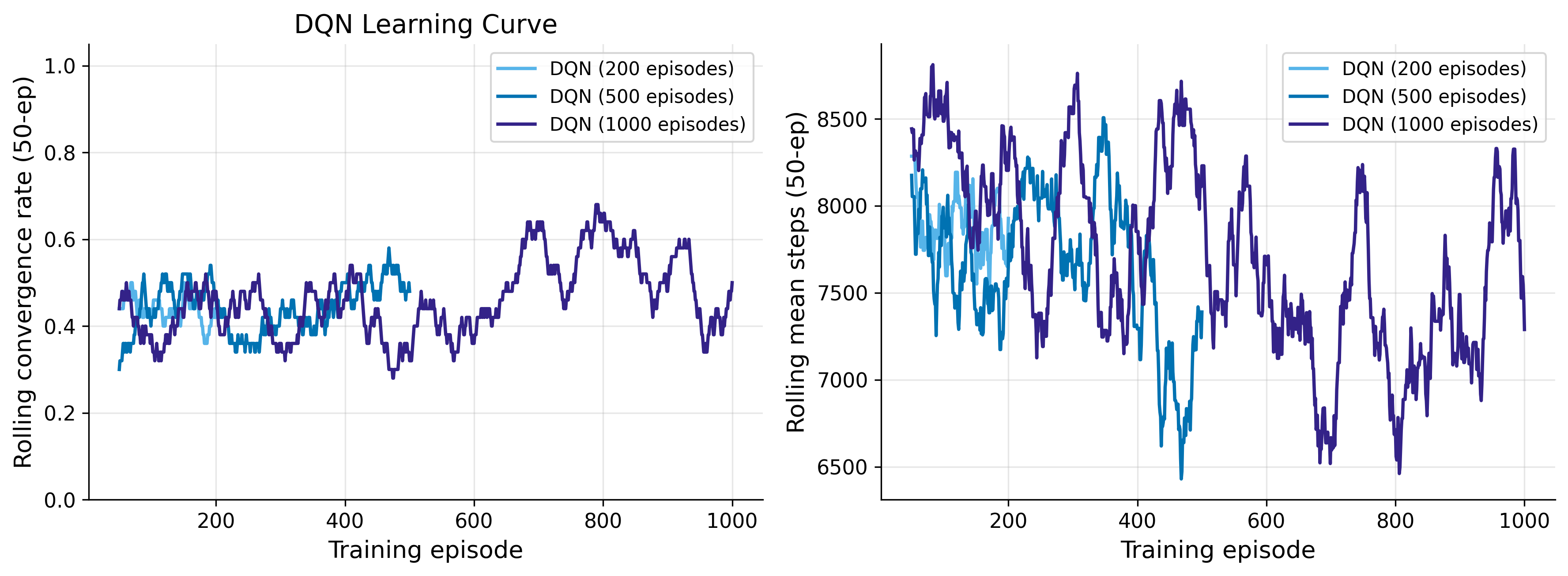}
    \end{minipage}
    \caption{\textbf{Left:} CWM vs.\ DQN on \jump{$_k$}: 100\% success rate vs.\ 58\%. \textbf{Right:} DQN learning curves ($k{=}2$). Training beyond 500 episodes degrades performance---success rate plateaus at ${\sim}50\%$ while steps decrease only marginally, revealing overfitting to the $\epsilon$-greedy exploration policy.}
    \label{fig:dqn_comparison}
\end{figure}

DQN achieves only 58\% success rate with 4$\times$ more steps than CWM (Figure~\ref{fig:dqn_comparison}, left). Notably, training longer \emph{hurts}: DQN with 1{,}000 episodes achieves ${\sim}60\%$ rolling success rate during training (Figure~\ref{fig:dqn_comparison}, right) but drops to 0\% when evaluated greedily---the Q-network overfits to the $\epsilon$-greedy exploration noise, learning to exploit frequent transitions rather than the rare but critical valley crossing. Without exploration noise at evaluation time, this overfitted policy never selects $k = k_\text{jump}$ at the valley edge.
In contrast, the CWM encodes the \emph{structure} of the problem into code, making it immune to this exploration--exploitation mismatch: the LLM extracts mathematical relationships from 200 offline trajectories that gradient-based RL cannot discover from 1{,}000 online episodes.
A DQN with richer state features or curriculum learning might narrow the gap, though the fundamental difficulty of discovering rare valley-crossing transitions via $\epsilon$-greedy exploration remains.

\subsection{NK-Landscape: Beyond Mathematical Models}
\label{sec:results_nk}

The previous benchmarks all have known mathematical structure that the LLM can exploit. The NK-Landscape~\cite{kauffman1987nk} removes this advantage: its fitness $f_{\text{NK}}(x) = \sum_i c_i(x_i, x_{N_1(i)}, \ldots, x_{N_K(i)})$ depends on random epistatic interactions\footnote{In the NK model, \emph{epistatic neighbours} of bit $i$ are the $K$ other bit positions whose values jointly determine the fitness contribution $c_i$. Higher $K$ introduces more inter-variable dependencies, increasing landscape ruggedness.} that produce a tunably rugged landscape with no closed-form transition model. This makes NK the hardest test for CWM---the model must learn entirely from data.

Since no mathematical description can be given to the LLM, the synthesis prompt relies solely on an \emph{empirical transition table}: for each (fitness range, $k$) pair, the table reports $P(\text{improve})$ and mean $\Delta f$, computed from the 300 training trajectories (as in \jump{$_k$}, Section~\ref{sec:results_jk}).
Figure~\ref{fig:nk_prompt_cwm} illustrates the synthesis pipeline for NK: the LLM receives the empirical table (left) and produces a Python CWM that encodes it as a lookup (right).

\begin{figure}[t]
\centering
\begin{minipage}[t]{0.48\textwidth}
\centering
\textbf{(A) Prompt excerpt (to LLM)}\vspace{2pt}
\begin{lstlisting}[style=promptstyle]
|{\color{promptkey}\bfseries \#\# Empirical Transition Table}|
|{\color{codepurple}fitness     k  P(impr)  mean\_df}|
[20, 25)    1   |{\color{promptnum}0.508}|   |{\color{promptnum}+0.307}|
[20, 25)   25   |{\color{promptnum}0.865}|   |{\color{promptnum}+2.219}|
[30, 32)    1   |{\color{promptnum}0.166}|   |{\color{promptnum}+0.083}|
[30, 32)    5   |{\color{promptnum}0.065}|   |{\color{promptnum}+0.046}|
[34, 36)    1   |{\color{promptnum}0.002}|   |{\color{promptnum}+0.001}|
[34, 36)    2   |{\color{promptnum}0.033}|   |{\color{promptnum}+0.012}|
[36, 38)    1   |{\color{promptnum}0.001}|   |{\color{promptnum}+0.000}|
|{\color{codegray}...}|
|{\color{promptcomment}Use this table in predict\_next\_state.}|
|{\color{promptcomment}At low fitness, large k helps.}|
|{\color{promptcomment}At high fitness, only small k works.}|
\end{lstlisting}
\end{minipage}%
\hfill
\begin{minipage}[t]{0.48\textwidth}
\centering
\textbf{(B) Synthesized CWM (by LLM)}\vspace{2pt}
\begin{lstlisting}[style=pythonstyle]
class |{\color{codepurple}SynthesizedCWM}|:
  T = { |{\color{codegreen}\# (fitness): \{k: (P, df)\}}|
    (20,25): {1:|{\color{codenumber}(.51,.31)}|, 25:|{\color{codenumber}(.87,2.2)}|},
    (30,32): {1:|{\color{codenumber}(.17,.08)}|,  5:|{\color{codenumber}(.07,.05)}|},
    (34,36): {1:|{\color{codenumber}(.00,.00)}|,  2:|{\color{codenumber}(.03,.01)}|},
    |{\color{codegray}...}| }

  def |{\color{codepurple}predict}|(self, state, action):
    p, df = self.T[bin(state)][action.k]
    new_f = state.fitness + df
    return |{\color{codepurple}NKState}|(fitness=new_f, |{\color{codegray}...}|)

  def |{\color{codepurple}evaluate}|(self, state):
    return state.normalized_fitness

  def |{\color{codepurple}actions}|(self, state):
    if state.stagnation > |{\color{codenumber}100}|:
      return [k=|{\color{codenumber}5}|, k=|{\color{codenumber}10}|, k=|{\color{codenumber}25}|]
    return [k=|{\color{codenumber}1}|, k=|{\color{codenumber}2}|, k=|{\color{codenumber}5}|]
\end{lstlisting}
\end{minipage}
\caption{NK-Landscape synthesis (simplified). \textbf{(A)}~The prompt provides an empirical transition table instead of a mathematical model. \textbf{(B)}~The LLM encodes this table into a Python CWM; the greedy planner selects $k^* = \arg\max_k \texttt{evaluate}(\texttt{predict}(s, k))$ at each step.}
\label{fig:nk_prompt_cwm}
\end{figure}

Table~\ref{tab:nk_results} presents results ($n=50$, $K=2$) aggregated across fifteen independently generated NK-Landscape instances with 100 runs per instance (1{,}500 total runs per policy).
CWM-greedy achieves the highest mean best fitness (36.94), outperforming all baselines including \texttt{static\_5} (36.32) and \texttt{random\_k} (36.31). Notably, CWM-greedy ranks first on \emph{every} instance individually---with margins ranging from $+0.41$ to $+0.99$---not just in aggregate (paired $t$-test $p < 0.001$ against every baseline).
The CWM was synthesized with mean Kendall's $\tau = 0.553$---substantially better than without the transition table ($\tau = -0.137$), confirming that structured empirical summaries can substitute for closed-form models.

\begin{table}[t]
\centering
\caption{NK-Landscape results ($n=50$, $K=2$, budget${}=10{,}000$, 100 runs $\times$ 15 instances). Std: across per-instance means; Max: best single run. CWM-greedy ranks first on every instance (paired $t$-test $p<0.001$).}
\label{tab:nk_results}
\small
\begin{tabular}{@{}lrrrc@{}}
\toprule
\textbf{Policy} & \textbf{Mean best} & \textbf{Std} & \textbf{Max} & \textbf{Mean rank} \\
\midrule
\textbf{CWM-greedy}        & \textbf{36.94} & \textbf{1.03} & \textbf{39.12} & \textbf{1.0} \\
\texttt{static\_5}         & 36.32 & 1.04 & 38.96 & 2.5 \\
\texttt{random\_k}         & 36.31 & 1.00 & 39.06 & 2.5 \\
\texttt{fifth\_rule}       & 35.36 & 0.99 & 38.91 & 4.5 \\
\texttt{stagnation}        & 35.32 & 0.93 & 38.91 & 5.1 \\
\texttt{self\_adjusting}   & 35.22 & 0.95 & 39.06 & 6.1 \\
\texttt{static\_1}         & 35.21 & 0.97 & 38.98 & 6.3 \\
\texttt{adaptive\_fitness} & 34.16 & 1.12 & 37.90 & 8.0 \\
\bottomrule
\end{tabular}
\end{table}

\begin{figure}[t]
    \centering
    \includegraphics[width=0.65\textwidth]{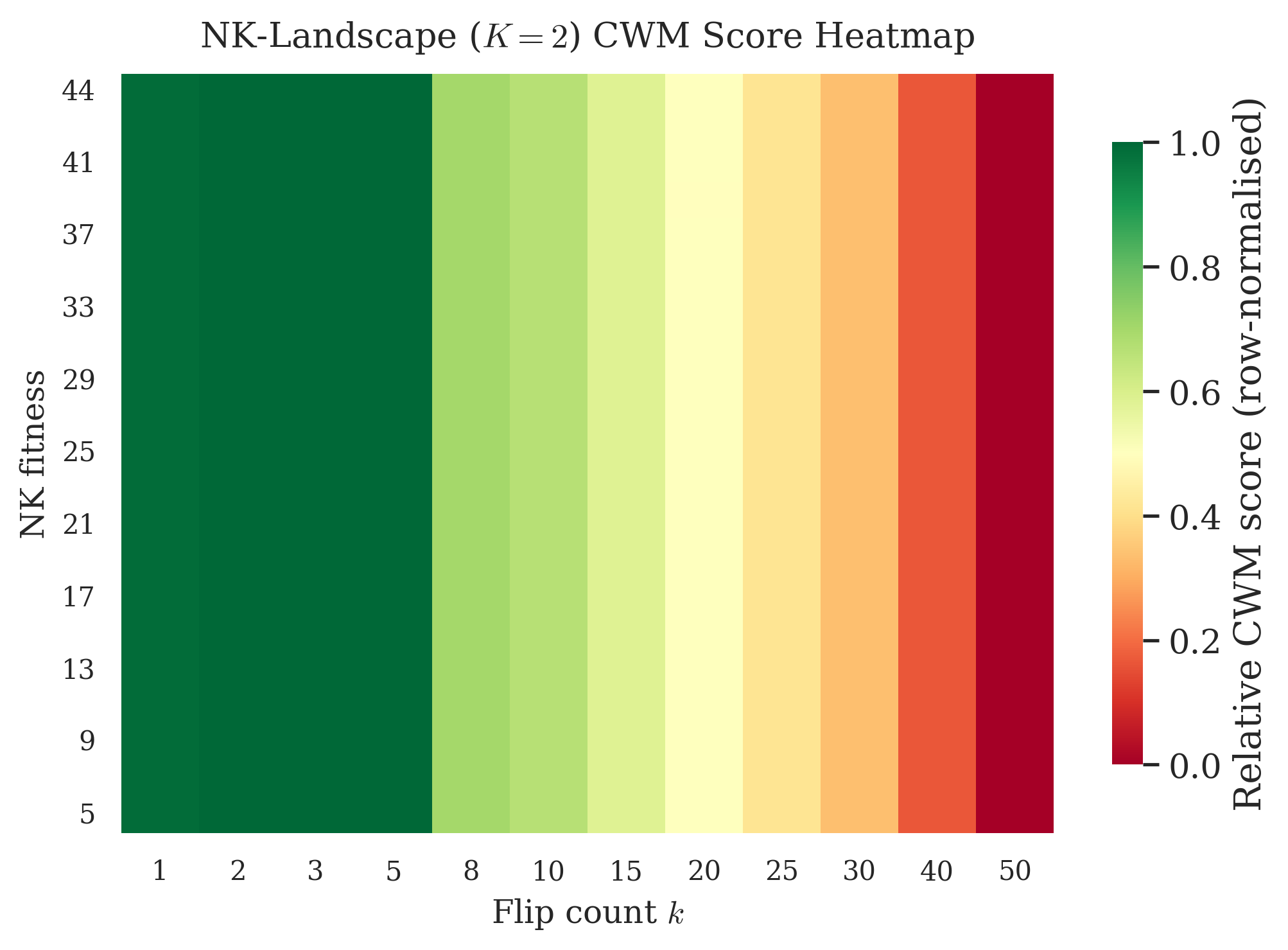}
    \caption{NK CWM score heatmap ($K{=}2$). The scores are instance-independent (determined by the CWM alone); the model is evaluated on fifteen landscape instances in Table~\ref{tab:nk_results}. No optimal overlay exists---the optimal policy is unknown. The CWM prefers small $k$ at high fitness.}
    \label{fig:nk_heatmap}
\end{figure}

Figure~\ref{fig:nk_heatmap} shows the CWM's score landscape: at high fitness, the model correctly favors small $k$, consistent with the diminishing returns of large mutations near local optima.

To test whether the CWM generalizes to more rugged landscapes without re-synthesis, we evaluate the same model (trained on $K{=}2$) on $K{=}3$ and $K{=}4$.
CWM-greedy achieves the best performance at $K{=}3$ ($36.93$ vs.\ $36.21$ for \texttt{static\_5}) and $K{=}4$ ($37.42$ vs.\ $36.57$), outperforming all baselines across all $K$ values.

\subsection{Generalization}
\label{sec:results_gen}

\paragraph{Across problem sizes.}
Table~\ref{tab:gen_size} shows that CWM-greedy, trained on $n{=}50$, maintains its advantage on $n{=}100$ and $n{=}200$ for both \lo{} and \onemax{}.
On \lo{}, CWM-greedy maintains a ratio of 1.02--1.12$\times$ optimal across all sizes, while RLS$_1$ degrades to 1.28$\times$---the CWM's structural model scales better than fixed policies.
On \onemax{}, CWM-greedy performs within 2\% of optimal across all sizes; at $n{=}200$ the CWM takes slightly fewer steps (1{,}020 vs.\ 1{,}031), though this difference is within noise.

\begin{table}[t]
\centering
\caption{Generalization across problem sizes. CWM trained on $n{=}50$, 50 eval.\ runs.}
\label{tab:gen_size}
\small
\begin{tabular}{@{}llrrrr@{}}
\toprule
\textbf{Problem} & $n$ & \textbf{Optimal} & \textbf{CWM} & \textbf{RLS$_1$} & \textbf{CWM/Opt} \\
\midrule
\multirow{3}{*}{\lo{}} & 50  &    983 & 1{,}052 & 1{,}220 & 1.07$\times$ \\
                       & 100 & 3{,}705 & 4{,}167 & 4{,}923 & 1.12$\times$ \\
                        & 200 & 15{,}761 & 16{,}156 & 20{,}210 & 1.02$\times$ \\
\midrule
\multirow{3}{*}{\onemax{}} & 50  & 185 & 188 & 192 & 1.02$\times$ \\
                            & 100 & 441 & 452 & 431 & 1.02$\times$ \\
                           & 200 & 1{,}031 & 1{,}020 & 1{,}037 & 0.99$\times$ \\
\bottomrule
\end{tabular}
\end{table}

\paragraph{Across Jump$_k$ values.}
Table~\ref{tab:gen_jump} presents the most striking generalization result: the CWM, trained on $k{=}2$, generalizes to $k{=}3$ with 78\% success rate---comparable to the greedy-optimal policy (88\%)---while DQN and $\EAalpha{}$ fail completely (0\%).
The budget is raised to $50{,}000$ for $k \geq 3$ to accommodate the larger expected waiting times ($\binom{50}{3} = 19{,}600$; $\binom{50}{4} \approx 230{,}000$).
Because the CWM encodes the parametric hypergeometric model (not just the empirical table), it correctly computes valley-crossing probabilities for unseen $k$ values.

\begin{table}[t]
\centering
\caption{Generalization across \jump{$_k$} values. CWM and DQN (best: 500 training episodes) trained on $k{=}2$, 50 eval.\ runs.}
\label{tab:gen_jump}
\small
\begin{tabular}{@{}lrrrrrr@{}}
\toprule
& \multicolumn{2}{c}{\textbf{CWM}} & \multicolumn{2}{c}{\textbf{DQN}} & \multicolumn{2}{c}{$\EAalpha$} \\
\cmidrule(lr){2-3} \cmidrule(lr){4-5} \cmidrule(lr){6-7}
$k$ & Mean & SR & Mean & SR & Mean & SR \\
\midrule
2 & 1{,}259 & 100\% & 5{,}023 & 58\% & 10{,}000 & 0\% \\
3 & 21{,}377 & 78\% & 50{,}000 & 0\% & 50{,}000 & 0\% \\
4 & 50{,}000 & 0\% & 50{,}000 & 0\% & 50{,}000 & 0\% \\
\bottomrule
\end{tabular}
\end{table}

\begin{figure}[t]
    \centering
    \includegraphics[width=0.65\textwidth]{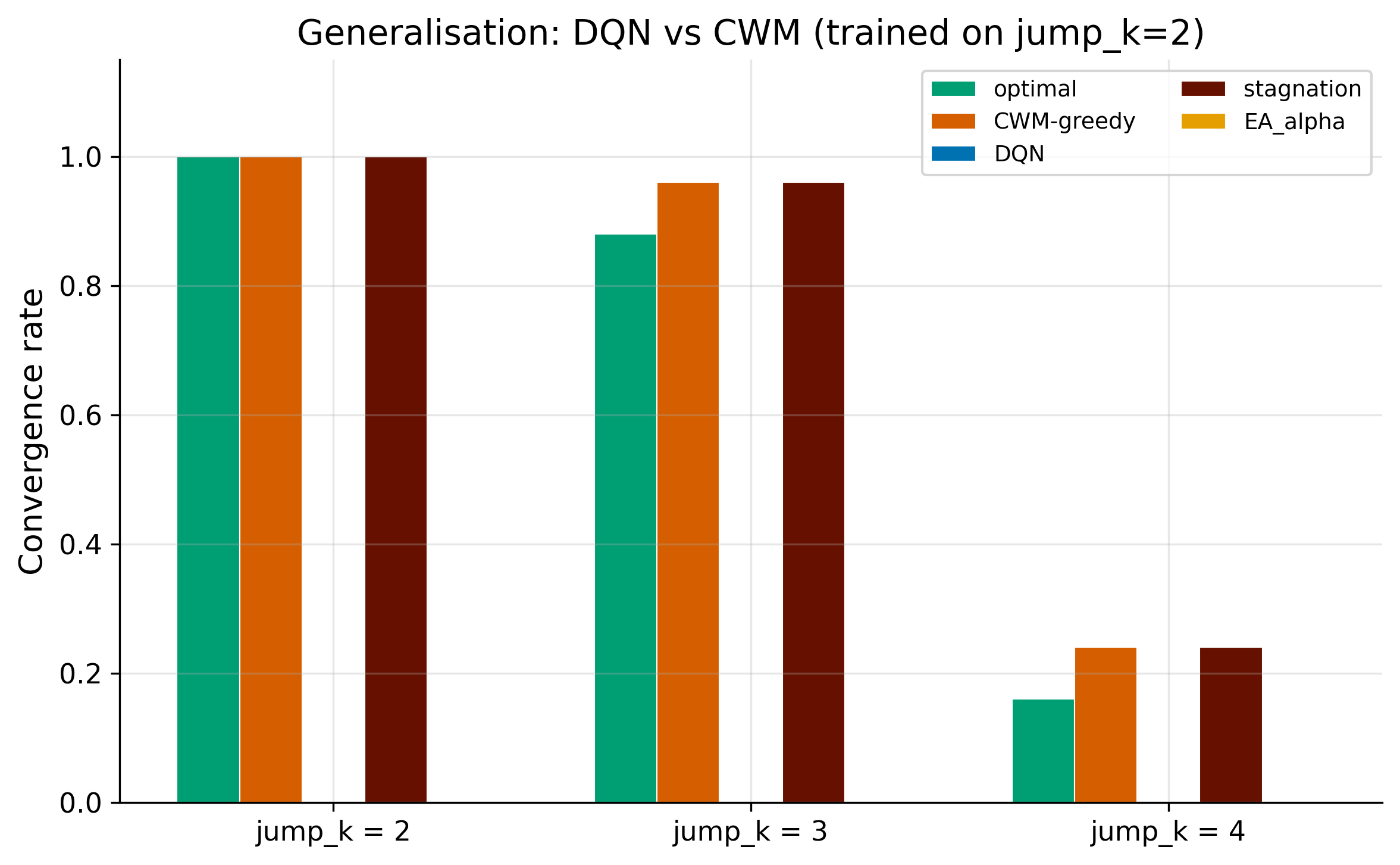}
    \caption{\jump{$_k$} generalization: CWM 78\% on $k{=}3$ vs.\ DQN 0\%.}
    \label{fig:gen_jump}
\end{figure}

Figure~\ref{fig:gen_jump} visualizes this gap: CWM maintains high success rates as $k$ grows, while DQN drops from 58\% ($k{=}2$) to 0\% ($k{\geq}3$) and $\EAalpha{}$ fails across all values.

\subsection{CWM Quality}
\label{sec:results_quality}

Table~\ref{tab:cwm_quality} summarizes CWM accuracy across benchmarks.
The \jump{$_k$} CWM achieves the highest correlation: the combined math + empirical prompt enables the LLM to synthesize a hypergeometric transition model that ranks actions correctly across all fitness levels. Valley-edge $\tau = 1.0$ confirms perfect ranking at the critical decision point.

\begin{table}[t]
\centering
\caption{CWM quality metrics. $\tau$: Kendall's rank correlation between CWM-predicted and true action rankings. $\rho$: Spearman correlation. Exact: fraction of states where CWM selects the same $k^*$ as the oracle.}
\label{tab:cwm_quality}
\small
\begin{tabular}{@{}lcccc@{}}
\toprule
\textbf{Benchmark} & $\tau$ & $\rho$ & \textbf{Exact} & \textbf{Note} \\
\midrule
\lo{}       & 0.784 & 0.916 & 60\% & --- \\
\onemax{}   & 0.614 & 0.939 & 82\% & --- \\
\jump{$_k$} & 0.898 & 0.948 & ---  & valley-edge $\tau{=}1.0$ \\
NK          & 0.553 & ---   & ---  & with empirical table \\
\bottomrule
\end{tabular}
\end{table}

\section{Discussion}
\label{sec:discussion}

\paragraph{Adaptive baselines and stagnation heuristic on Jump$_k$.}
On \jump{$_k$}, $\EAalpha{}$ and self-adjusting mechanisms decrease $k$ during stagnation---a consequence of their multiplicative update rules designed for unimodal functions. The valley requires the \emph{opposite}: increasing $k$. No $(A, b)$ configuration can fix this structural mismatch.
The \texttt{stagnation} heuristic achieves 99\% success rate (1{,}506 steps vs.\ CWM's 1{,}342), but requires $k_\text{jump}$ as a hard-coded parameter. The CWM \emph{infers} this strategy from the mathematical description enriched with empirical transition statistics---without any collection policy using $k_\text{jump}$---and generalizes to $k{=}3$ (78\% SR) without reconfiguration.
An alternative is \emph{heavy-tailed mutation}~\cite{doerr2017fast,friedrich2018heavy}, which handles Jump$_k$ without knowing $k$ via a fixed power-law distribution; a direct comparison is future work.

\paragraph{Does the prompt give the answer?}
A natural concern is that encoding the fitness function in the prompt trivializes the task. We argue it does not, for three reasons.
First, this mirrors standard practice: the original CWM framework~\cite{dainese2024cwm} similarly provides environment documentation to the LLM.
Second, knowing the mathematics is not enough---the LLM must convert it into correct probabilistic code, a non-trivial step that requires up to 5 validation attempts.
Third, and most importantly, on \jump{$_k$} and NK the CWM succeeds with \emph{empirical} transition statistics instead of mathematical descriptions (NK: $\tau = 0.553$ vs.\ $-0.137$ without the table), showing that structured data can substitute for formal models.
Furthermore, the training trajectories contain no oracle policies: no collection policy uses the optimal $k^*(i)$ or, for \jump{$_k$}, even the value of $k_\text{jump}$. The key input is \emph{structured transition information}---not imitation of optimal demonstrations.

\paragraph{Robustness.}
Five independent syntheses (temperatures 0.6--1.0) yield stable results: LO $1020 \pm 20$ steps ($\rho = 0.916$); OM $190 \pm 5$ steps ($\rho = 0.939$); Jump$_k$ $1342$ steps, 100\% success rate, valley-edge $\tau = 1.0$; NK $37.76 \pm 0.2$ best fitness ($\tau = 0.553$)---critical behavior is robustly captured across independent synthesis runs.
For NK, we additionally evaluate across fifteen independently generated landscape instances: CWM-greedy ranks first on every instance (paired $t$-test $p < 0.001$ vs.\ all baselines), confirming that the synthesized policy generalizes beyond the training landscape.

\paragraph{Limitations.}
(1)~For structured benchmarks, the prompt includes the mathematical model; for \jump{$_k$} and NK, the prompt is enriched with empirical transition statistics computed from training data.
(2)~On OneMax, CWM-greedy is indistinguishable from \texttt{RLS\_1}.
(3)~On NK, CWM requires \emph{structured} transition information; with only a qualitative description it loses to baselines.
(4)~For $k{=}4$, CWM achieves 0\% success rate ($\binom{50}{4} \approx 230{,}000 \gg$ budget); even the greedy-optimal policy reaches only 16\%.
(5)~NK results are limited to $K \in \{2,3,4\}$.

\section{Conclusion}
\label{sec:conclusion}

We have shown that Code World Models---originally designed for deterministic games---can be extended to stochastic combinatorial optimization. From suboptimal trajectories alone, the LLM synthesizes a simulator accurate enough for greedy planning to match or exceed all adaptive baselines, including 100\% success rate on the deceptive \jump{$_k$} landscape where every other adaptive method fails. On NK, where no mathematical model exists, empirical transition statistics suffice. Across all benchmarks, the CWM outperforms DQN while requiring less data and producing an auditable Python program instead of opaque network weights.

Code and data are available at \url{https://github.com/camilochs/cwm-adaptive-control}.

Future work will extend CWMs to continuous optimization, multi-parameter control, population-based algorithms, NK with higher $K$ values, and comparison with heavy-tailed mutation operators~\cite{doerr2017fast,friedrich2018heavy}.

A broader point deserves emphasis. As generative AI and human-authored code increasingly coexist~\cite{georgiev2025mathematicalexplorationdiscoveryscale,sartori2025architectureserrorphilosophicalinquiry}, a methodological question arises: can the probabilistic, black-box nature of an LLM be reconciled with a field built on asymptotic guarantees? The CWM framework offers one answer: by forcing the LLM to express its knowledge as auditable code, it translates statistical experience into an explicit heuristic---complementing rather than supplanting classical theory, and enabling analytical planning even in landscapes devoid of closed-form models.


\bibliographystyle{splncs04}
\bibliography{references}

@misc{sartori2025architectureserrorphilosophicalinquiry,
  title        = {{Architectures of Error: A Philosophical Inquiry into AI and Human Code Generation}},
  author       = {Chacón Sartori, Camilo},
  year         = {2025},
  note         = {In press at Philosophy \& Technology},
  eprint       = {2505.19353},
  archivePrefix= {arXiv},
  primaryClass = {cs.AI},
  url          = {https://arxiv.org/abs/2505.19353}
}

@article{aleti2016systematic,
  title={A systematic literature review of adaptive parameter control methods for evolutionary algorithms},
  author={Aleti, Aldeida and Moser, Irene},
  journal={ACM Computing Surveys (CSUR)},
  volume={49},
  number={3},
  pages={1--35},
  year={2016},
  publisher={ACM}
}

@inproceedings{dainese2024cwm,
  title={Generating code world models with large language models guided by {Monte Carlo Tree Search}},
  author={Dainese, Nicola and Merler, Michele and Alakuijala, Matti and Marttinen, Pekka},
  booktitle={Advances in Neural Information Processing Systems},
  year={2024}
}

@inproceedings{doerr2018ea_alpha,
  title={On the effectiveness of simple success-based parameter selection mechanisms for two classical discrete black-box optimization benchmark problems},
  author={Doerr, Carola and Wagner, Markus},
  booktitle={Proceedings of the Genetic and Evolutionary Computation Conference},
  pages={943--950},
  year={2018},
  organization={ACM}
}

@inproceedings{doerr2019self_adjusting,
  title={Self-adjusting mutation rates with provably optimal success rules},
  author={Doerr, Benjamin and Doerr, Carola and Lengler, Johannes},
  booktitle={Proceedings of the Genetic and Evolutionary Computation Conference},
  pages={1479--1487},
  year={2019},
  organization={ACM}
}

@inproceedings{doerr2017fast,
  title={Fast genetic algorithms},
  author={Doerr, Benjamin and Le, Huu Phuoc and Makhmara, R{\'e}gis and Nguyen, Ta Duy},
  booktitle={Proceedings of the Genetic and Evolutionary Computation Conference},
  pages={777--784},
  year={2017},
  organization={ACM}
}

@article{droste2002analysis,
  title={On the analysis of the (1+1) evolutionary algorithm},
  author={Droste, Stefan and Jansen, Thomas and Wegener, Ingo},
  journal={Theoretical Computer Science},
  volume={276},
  number={1--2},
  pages={51--81},
  year={2002},
  publisher={Elsevier}
}

@inproceedings{friedrich2018heavy,
  title={Heavy-tailed mutation operators in single-objective combinatorial optimization},
  author={Friedrich, Tobias and G{\"o}bel, Andreas and Quinzan, Francesco and Wagner, Markus},
  booktitle={Parallel Problem Solving from Nature -- PPSN XV},
  series={Lecture Notes in Computer Science},
  volume={11102},
  pages={134--145},
  year={2018},
  organization={Springer}
}

@article{ha2018world,
  title={World models},
  author={Ha, David and Schmidhuber, J{\"u}rgen},
  journal={arXiv preprint arXiv:1803.10122},
  year={2018}
}

@article{jansen2002analysis,
  title={The analysis of evolutionary algorithms---a proof that crossover really can help},
  author={Jansen, Thomas and Wegener, Ingo},
  journal={Algorithmica},
  volume={34},
  number={1},
  pages={47--66},
  year={2002},
  publisher={Springer}
}

@article{kauffman1987nk,
  title={Towards a general theory of adaptive walks on rugged landscapes},
  author={Kauffman, Stuart A. and Levin, Simon},
  journal={Journal of Theoretical Biology},
  volume={128},
  number={1},
  pages={11--45},
  year={1987},
  publisher={Elsevier}
}

@article{karafotias2015parameter,
  title={Parameter control in evolutionary algorithms: Trends and challenges},
  author={Karafotias, Giorgos and Hoogendoorn, Mark and Eiben, {\'A}goston E.},
  journal={IEEE Transactions on Evolutionary Computation},
  volume={19},
  number={2},
  pages={167--187},
  year={2015},
  publisher={IEEE}
}

@incollection{eiben2007rl_ea,
  title={Reinforcement learning for online control of evolutionary algorithms},
  author={Eiben, {\'A}goston E. and Horvath, M. and Kowalczyk, Wojtek and Schut, Martijn C.},
  booktitle={Engineering Self-Organising Systems},
  series={Lecture Notes in Computer Science},
  volume={4335},
  pages={151--160},
  year={2007},
  publisher={Springer}
}

@article{mnih2015human,
  title={Human-level control through deep reinforcement learning},
  author={Mnih, Volodymyr and Kavukcuoglu, Koray and Silver, David and Rusu, Andrei A. and Veness, Joel and Bellemare, Marc G. and Graves, Alex and Riedmiller, Martin and Fidjeland, Andreas K. and Ostrovski, Georg and others},
  journal={Nature},
  volume={518},
  number={7540},
  pages={529--533},
  year={2015},
  publisher={Nature Publishing Group}
}

@inproceedings{opris2024tight,
  title={A tight {${O(4^k/p_c)}$} runtime bound for a {${(\mu+1)}$} {GA} on {Jump$_k$} for realistic crossover probabilities},
  author={Opris, Andrei and Lengler, Johannes and Sudholt, Dirk},
  booktitle={Proceedings of the Genetic and Evolutionary Computation Conference},
  pages={1605--1613},
  year={2024},
  organization={ACM}
}

@book{rechenberg1973,
  title={Evolutionsstrategie: Optimierung technischer Systeme nach Prinzipien der biologischen Evolution},
  author={Rechenberg, Ingo},
  year={1973},
  publisher={Frommann-Holzboog}
}

@article{romera2024funsearch,
  title={Mathematical discoveries from program search with large language models},
  author={Romera-Paredes, Bernardino and Barekatain, Mohammadamin and Novikov, Alexander and Balog, Matej and Kumar, M. Pawan and Dupont, Emilien and Ruiz, Francisco J. R. and Stephens, Charles R. and Fawzi, Alhussein and others},
  journal={Nature},
  volume={625},
  number={7995},
  pages={468--475},
  year={2024},
  publisher={Nature Publishing Group}
}

@article{schrittwieser2020muzero,
  title={Mastering {Atari}, {Go}, chess and shogi by planning with a learned model},
  author={Schrittwieser, Julian and Antonoglou, Ioannis and Hubert, Thomas and Simonyan, Karen and Sifre, Laurent and Schmitt, Simon and Guez, Arthur and Lockhart, Edward and Hassabis, Demis and Graepel, Thore and Lillicrap, Timothy and Silver, David},
  journal={Nature},
  volume={588},
  number={7839},
  pages={604--609},
  year={2020},
  publisher={Nature Publishing Group}
}

@inproceedings{sharma2019deep,
  title={Deep reinforcement learning based parameter control in differential evolution},
  author={Sharma, Mudita and Komninos, Alexandros and L{\'o}pez-Ib{\'a}{\~n}ez, Manuel and Kazakov, Dimitar},
  booktitle={Proceedings of the Genetic and Evolutionary Computation Conference},
  pages={709--717},
  year={2019},
  organization={ACM}
}

@misc{georgiev2025mathematicalexplorationdiscoveryscale,
      title={Mathematical exploration and discovery at scale}, 
      author={Bogdan Georgiev and Javier Gómez-Serrano and Terence Tao and Adam Zsolt Wagner},
      year={2025},
      eprint={2511.02864},
      archivePrefix={arXiv},
      primaryClass={cs.NE},
      url={https://arxiv.org/abs/2511.02864}, 
}
\end{document}